%% file: main.tex
\documentclass[manuscript,screen,nonacm]{acmart}
\usepackage{amsmath,amsfonts}

\usepackage{algorithmic}
\usepackage{diagbox}
\usepackage{url}
\usepackage{caption}
\usepackage{subcaption}
\usepackage{booktabs}
\usepackage{hyperref}
\usepackage{color}
\usepackage{xcolor}
\usepackage{colortbl}
\usepackage[flushleft]{threeparttable}
\usepackage{tikz}
\usepackage{xspace}
\usepackage{bbding} 
\usepackage{romannum}
\usepackage{enumitem}
\usepackage[ruled,linesnumbered]{algorithm2e}

\newtheorem{remark}{Remark}

\input{tex/macros}

\AtBeginDocument{%
  }

\setcopyright{acmcopyright}
\copyrightyear{2022}
\acmYear{2022}
\acmDOI{XXXXXXX.XXXXXXX}

\acmPrice{15.00}
\acmISBN{978-1-4503-XXXX-X/18/06}

\begin{document}

\pagenumbering{arabic}

\title[Explaining with greater support: q-consistent summary explanation]{Explaining with Greater Support: Weighted Column Sampling Optimization for $q$-Consistent Summary-Explanations}

\input{tex/002authors}
\input{tex/001abstract}






\maketitle

\input{tex/010introduction}
\input{tex/020background}
\input{tex/030problem}
\input{tex/040methodology}
\input{tex/050experiments}
\input{tex/060conclusion}
\input{tex/003acknowledgement}



\bibliographystyle{ACM-Reference-Format}
\bibliography{main}

\end{document}

%% file: tex/macros.tex
\SetKwInOut{Parameter}{Hyper-parameter}
\SetKwIF{If}{ElseIf}{Else}{if}{}{else if}{else}{end if}%
\SetKwFor{While}{while}{}{end while}%
\SetKwRepeat{Do}{do}{while}
\SetKwInOut{Parameter}{parameter}
\SetKwIF{If}{ElseIf}{Else}{if}{}{else if}{else}{end if}%
\SetKwFor{While}{while}{}{end while}%
\SetKwRepeat{Do}{do}{while}
\SetKwInOut{Parameter}{Hyper-parameter}



%% file: tex/002authors.tex


\author{CHEN PENG}
\email{chen.peng@jsu.edu.cn}
\affiliation{
  \institution{Jishou University}
  \country{China}
}
\author{ZHENGQI DAI}
\affiliation{
  \institution{Jishou University}
  \country{China}
}
\author{GUANGPING XIA}
\affiliation{
  \institution{Jishou University}
  \country{China}
}
\author{YAJIE NIU}
\affiliation{
  \institution{Jishou University}
  \country{China}
}
\author{YIHUI LEI}
\affiliation{
  \institution{Jishou University}
  \country{China}
}

\renewcommand{\shortauthors}{Peng et al.}

%% file: tex/001abstract.tex
\begin{abstract}
Machine learning systems have been extensively used as auxiliary tools in domains that require critical decision-making, such as healthcare and criminal justice. 
The explainability of decisions is crucial for users to develop trust on these systems. 
In recent years, the globally-consistent rule-based summary-explanation and its max-support (MS) problem have been proposed, which can provide explanations for particular decisions along with useful statistics of the dataset. 
However, globally-consistent summary-explanations with limited complexity typically have small supports, if there are any. 
In this paper, we propose a relaxed version of summary-explanation, i.e., the $q$-consistent summary-explanation, which aims to achieve greater support at the cost of slightly lower consistency. 
The challenge is that the max-support problem of $q$-consistent summary-explanation (MSqC) is much more complex than the original MS problem, resulting in over-extended solution time using standard branch-and-bound solvers.
To improve the solution time efficiency, this paper proposes the weighted column sampling~(WCS) method based on solving smaller problems by sampling variables according to their simplified increase support (SIS) values. 
Experiments verify that solving MSqC with the proposed SIS-based WCS method is not only more scalable in efficiency, but also yields solutions with greater support and better global extrapolation effectiveness.


\end{abstract}

%% file: tex/010introduction.tex
\section{Introduction}\label{Sec:intro}

With the advances in machine learning and the explosion of artificial intelligence applications, the transparency of machine learning models in many contexts has become increasingly important.
Some domains that require critical decision making, such as healthcare and criminal justice, have begun incorporating machine learning systems as support~\cite{lawlessInterpretableFairBoolean2021, wexler2017computer, varshney2017safety, dong2021survey}.
However, models with high accuracies are often black boxes, which are opaque and have their decision logic hidden.
This is mostly unintentional and inevitable, because the model (or algorithm) is not directly coded by men, but is generated from a complex hypothesis set, e.g., a neural network or support vector machine with  numerous parameters, through machine learning with a large amount of training data.

To achieve better transparency, traditionally, global explanation techniques have been used to extract approximate interpretable models from black-box ML models to mimic their decision logic.
For example, classification rules can be extracted from neural networks~\cite{andrewsSurveyCritiqueTechniques1995, qiao2021learning}, and decision trees can be extracted from tree ensembles~\cite{haraMakingTreeEnsembles2018, liu2022credit}.
However, each of these traditional global explanation methods is usually dedicated to a certain type of models, such as neural networks and tree ensembles, and thus they are not model-agnostic.
Furthermore, for a very complex model and a large dataset, the extracted interpretable model may not approximate well.

It was later realized that, while a black-box model can be arbitrarily complex, in the neighborhood of a data instance, the decision boundary should be simple enough to be captured by an interpretable model.
This is what drives the emergence of local explanation methods, which aims at describing model behavior with respect to a particular input.
In fact, the ability to explain a particular input-output pair is crucial for social acceptance and trust on a model's adoption in many sensitive applications.

A method which is most closely related to this paper is the \textit{globally-consistent rule-based summary-explanation} of Rudin \textit{et al.}~\cite{rudinGloballyConsistentRuleBasedSummaryExplanations2019},
where IP optimization problems are solved to find the rule-based summary-explanation of a target observation with minimum complexity (the MC problem) or maximized support (the MS problem).
The support of a rule-based summary-explanation is the number of instances in the dataset that meets the rule's IF-condition.
The following is an example globally-consistent summary-explanation given by the MaxSupport algorithm for an observation of the FICO dataset~\cite{ExplainableMachineLearning}, where the support is 594: ``For all 594 people whose ExternalRiskEstimate $\leq 63$ and AverageMinFile $\leq 48$, all of them were predicted to default''.
Clearly, explanations with large supports help users develop trust on the explainer system.

However, due to the global consistency constraint, when applied on a large dataset, solving the MS problem often results in rules with high complexity or small support, if not infeasible.
For many practical scenarios, explanations with large supports have a great impact on users' acceptance, while small inconsistency may be tolerable.
Motivated by this assumption, in this paper, the problem of maximizing support with $q$-consistency (MSqC), $q\in (0,1]$, is considered, which can significantly increase the support of the explanation rule.
The following is an example 0.85-consistent summary-explanation with support 2000:
``Over 85\% of the 2000 people, whose ExternalRiskEstimate $\leq 63$ and AverageMinFile $\leq 48$, were predicted to default''.

Although the extension seems straightforward, it turns out that the problem of finding a $q$-consistent summary-explanation is much more complex than finding a globally-consistent one. The reason is that, to maximize the $q$-consistent support which can have up to $1-q$ fraction of the matched observations being inconsistent, all observations that have different outcomes than the target observation have to be considered in the MSqC formulation, which are not present in MS.
Furthermore, standard integer programming (IP) solvers with branch-and-bound~(B\&B) exact solution are not very efficient on MS.
For example, in~\cite{rudinGloballyConsistentRuleBasedSummaryExplanations2019} a time limit of 60 seconds has to be set to terminate solution, and in our reproduction, full solution of MS using the SCIP solver~\cite{GamrathEtal2020ZR} takes 101 seconds for dataset size $|\mathcal{N}|=1$K and 1852 seconds for $|\mathcal{N}|=10$K on average.
The performance is much worse for the more complicated MSqC, as shown in the experiment.
Therefore, this paper proposes a weighted column sampling (WCS) multiple B\&B framework based on the simplified increased support (SIS) of the variables in MSqC, which is significantly more efficient and scalable than standard IP solvers using singleton branch-and-bound with a set time limit.
The major contributions of this paper are summarized as follows:
\begin{enumerate}
	\item The problem of MSqC is formulated, which can yield summary-explanations with much greater supports at the cost of slightly lower consistencies.
	\item The SIS-based WCS method is proposed, which is much more efficient and scalable than standard branch-and-bound IP solvers for MSqC.
	\item The global extrapolation effectiveness of the $q$-consistent summary-explanation is also considered in this paper, which is improved using a global prior injection technique in the SIS-based WCS method.
\end{enumerate}

This paper is organized as follows.
Section~\ref{Background} introduces the background and related studies.
Section~\ref{Problem formulation} formally defines the globally-consistent summary-explanation and $q$-consistent summary-explanation, and formulates the MC, MS, and MSqC problems.
Then, Section~\ref{Methodology} introduces the methods used in this paper, including the SIS of variables in MSqC, the WCS multiple B\&B framework, the global prior injection technique for global extrapolation, as well as various alternative improvements and performance metrics used in the experiments.
In Section~\ref{Computer experiments}, the time efficiency and solution effectiveness of the proposed SIS-based WCS method on MSqC is compared with various benchmark methods, including using the SCIP solver~\cite{GamrathEtal2020ZR} on MC, MS and MSqC, as well as a random column sampling (RCS) method.
Finally, Section~\ref{conclusions} concludes the paper.

%% file: tex/020background.tex
\section{Background}\label{Background}

Early explainability literature typically studies model-specific global explanation methods, which focus on describing the global behavior of specific models. For example, decision trees~\cite{cravenExtractingTreestructuredRepresentations1995, krishnanExtractingDecisionTrees1999, chen2022evaluation} and classification rules~\cite{andrewsSurveyCritiqueTechniques1995, augastaReverseEngineeringNeural2012, johanssonAccuracyVsComprehensibility2004} have been adopted to globally explain neural networks.
Differing from the traditional approaches, model-agnostic local explanation methods have been more popular in recent years, which focus on explaining the reasons for specific observations or predictions.
The emergence of this trend is often credited to the seminal work of Ribeiro {\it et al}.~\cite{ribeiroWhyShouldTrust2016}, where the LIME algorithm was proposed.
It was realized that model-level (i.e., global) explanation is not universally applicable, since a black-box model can be arbitrarily complex; however, in the neighborhood of a data instance, the decision boundary should always be simple enough to be captured by an interpretable model.
Such an assumption holds true for almost all realistic models, which makes local explanations widely applicable.

The LIME algorithm belongs to the category of \textit{model-available-agnostic} methods, where the black-box model is assumed to be available for local data generation (local to the target observation), though unknown.
Specifically, for any arbitrary classifier, the LIME algorithm~\cite{ribeiroWhyShouldTrust2016} first generates perturbed data around the target observation, then fits an interpretable model (a sparse linear model) on the generated local dataset for the model-agnostic local explanation.
Extensions have been made in follow-up studies.
For example, in a subsequent work~\cite{ribeiroAnchorsHighPrecisionModelAgnostic2018}, Ribeiro {\it et al}. proposed to replace the sparse linear model in LIME with if-then-rules, called Anchors, for more intuitive and accurate explanations.
In~\cite{guidottiLocalRuleBasedExplanations2018}, Guidotti~{\it et al}. used a genetic algorithm to generate the local dataset, and derived both decision rules and counterfactual rules from the trained local interpretable predictor. 
Zafar~\textit{et. al.}~\cite{zafar2021deterministic} proposed Deterministic Local Interpretable Model-Agnostic Explanations (DLIME) to overcome the data shift and instability caused by random perturbations in LIME. Huang~\textit{et. al.}~\cite{huang2022graphlime} created a general GNN model interpretation framework, GraphLIME, that provides a local interpretation of model results based on nonlinear feature selection such as Hilbert-Schmidt Independence Criterion (HSIC) Lasso.

There are cases where the black-box model is not available for synthetic data generation and the only knowledge source is the historical or provided data.
A well-known such scenario is the FICO explainable machine learning challenge~\cite{ExplainableMachineLearning}, where a dataset generated by the FICO company's black-box model is available, but the model itself is not accessible to researchers.
In such cases where only data is available, a common practice for explanation is to train another predictive model based on the provided dataset to approximate the original model.
Two options are available: the approximation model can either be an inherently interpretable model, or focus on accuracy by selecting a black box model for approximation.
\begin{enumerate}
	\item If the approximation model is chosen to be an inherently interpretable model, such as boolean rules or logistic regression model, then the interpretable model can be used for global explanation.
For example, in Dash~\textit{et. al.}~\cite{dashBooleanDecisionRules2018,   lawlessInterpretableFairBoolean2021}, boolean decision rules are generated using an approximate column generation algorithm. Wang~\textit{et. al.}~\cite{wang2021scalable} proposed the Rule-based Representation Learner (RRL) which automatically learns interpretable non-fuzzy rules for data representation and classification. 
In Wei~\textit{et. al.}~\cite{weiGeneralizedLinearRule2019}, a logistic regression model is fitted using also the column generation method.
However, it is hard to guarantee that an inherently interpretable model can closely approximate the original model.
	\item In the other way, if accuracy is prioritized, then a black box model such as a neural network can be used for approximation. Hence model-available-agnostic methods can be used for local explanation~\cite{ribeiroAnchorsHighPrecisionModelAgnostic2018, xia2018novel}.
	However, any interpretation is performed on the local data generated by the approximation model, instead of directly using the provided data.
\end{enumerate}
In summary, both ways require training a secondary model to approximate the original model, and the approximation error has to be sufficiently small for explanations to be trust-worthy.
Furthermore, either way cannot provide explanations that contain statistics relative to the provided dataset, such as the support information in~\cite{rudinGloballyConsistentRuleBasedSummaryExplanations2019}.

Following another path, Rudin \textit{et al.}~\cite{rudinGloballyConsistentRuleBasedSummaryExplanations2019} proposed the \textit{globally-consistent rule-based summary-explanation} method, which does not rely on generating local data using an available model.
Instead, the rule-based summary-explanation is found by solving an integer programming (IP) optimization problem, which is solely based on the dataset, and can be aimed at minimizing rule complexity or maximizing support.
The support of a rule-based summary-explanation is the number of instances in the dataset that meet the rule's IF-condition.
An example summary-explanation statement given by this approach for an observation of the FICO dataset is as follows: ``For all 594 people whose ExternalRiskEstimate $\leq 63$ and AverageMinFile $\leq 48$, all of them were predicted to default''.
The support of this summary-explanation is 594, which is useful information that the model-available-agnostic methods relying on data generation cannot compute.

A useful property of the summary-explanations produced by Rudin \textit{et al.}'s method~\cite{rudinGloballyConsistentRuleBasedSummaryExplanations2019} is that the explanations are guaranteed to be globally-consistent, i.e., for all instances in the dataset that meet the IF-condition, the THEN-statement is true.
However, globally-consistent rules can be difficult to find. 
When applied to practical large datasets, the approach of~\cite{rudinGloballyConsistentRuleBasedSummaryExplanations2019} often results in rules with high complexity or small support, if not infeasible.
In fact, for many practical scenarios, explanations with large supports help users develop trust on the explainer system, while small inconsistency may be tolerable (under certain threholds). Motivated by this idea, in this paper, the MaxSupport problem of~\cite{rudinGloballyConsistentRuleBasedSummaryExplanations2019} is extended by allowing small inconsistencies in the explanation, which effectively increases the support of the explanation rule.

%% file: tex/030problem.tex
\section{Problem formulation}\label{Problem formulation}
\subsection{Globally-Consistent Summary-Explanation}
Consider a $|\mathcal{P}|$-dimensional binary observation dataset $\{(x_i, y_i), i\in \mathcal{N}\}$, where $x_i\in \{0,1\}^{|\mathcal{P}|}$, $\mathcal{N}$ is the observation index set, and $\mathcal{P}$ is the index set of binary-valued feature functions.
In general, such a binary dataset can be derived from any dataset $\{(\tilde{x}_i, y_i), i\in \mathcal{N}\}$ with arbitrary inputs $\tilde{x}_i$. For example, a raw input $\tilde{x}_e=\{\tilde{x}_{e,1}, \tilde{x}_{e,2}\} = \{30, 10\}$ can be binarized as $x_e=\{\delta_{e,1}, \delta_{e,2}, \delta_{e,3}\} = \{1,0, 1\}$ using the ordered feature functions set $\{\tilde{x}_{e,1}\geq 0, \tilde{x}_{e,1}\geq 50, \tilde{x}_{e,2}\geq 0\}$. In the following, the terms ``feature'' and ``feature function'' are used synonymously.

Let $b$ denote a conjunctive clause, which is made of the logical AND ($\land$) operation of multiple conditions, each corresponding to a binary feature $F_p$, i.e., $b(\cdot)=\land_{p\in\mathcal{P}'}F_p(\cdot)$ for a subset of features $\mathcal{P}'\subseteq \mathcal{P}$.
For the above example, $b$ can be $\tilde{x}_{e,1}\geq 0$, or $(\tilde{x}_{e,1}\geq 50)\land (\tilde{x}_{e,2}\geq 0)$.

A \textit{summary-explanation} is a rule $b(\cdot)\to y$ that describes the binary classifier
$$
h^{b(\cdot)\rightarrow y}(x)=\begin{cases}
	y & \text{if~} b(x)=\land_{p\in\mathcal{P}'}F_p(x)=1,\\
	1-y & \text{otherwise}.
\end{cases}
$$
A \textit{globally-consistent summary-explanation} for an observation $(x_e, y_e)$ is a summary-explanation $b(\cdot)\to y_e$ with the following properties:
\begin{enumerate}
	\item \textit{relevancy}, i.e., $b(x_e)=1$;
	\item \textit{consistency}, i.e., for all observations $i\in\mathcal{N}$, if $b(x_i)=1$, then $y_i=y_e$.
\end{enumerate}
In a more readable language, a globally-consistent summary-explanation $b(\cdot)\rightarrow y_e$ can be expressed as: ``for all observations (e.g., people/customers) where $b(\cdot)$ is TRUE, the outcomes (e.g., predicted risk/decision) are $y_e$ (the same as observation $e$)''.
Such an explanation aligns the current observation $(x_e,y_e)$ with existing records in the dataset, thus is more convincing to users in fields like credit evaluation.

Note that a general global (model-level) rule-based explanation typically would use a rule list in disjunctive normal form (DNF) or conjunctive normal form (CNF)~\cite{dashBooleanDecisionRules2018}.
Suppose that a DNF rule list $b^{(1)}(\cdot)\lor b^{(2)}(\cdot)\lor \dots$ is a global explanation for the black box, then a globally-consistent summary-explanation for a specific observation $(x_e, y_e)$ can be viewed as one of the clauses $b(\cdot)$ in the DNF global rule list that is activated by $x_e$, i.e., $b(x_e)=1$.

The quality of the clause $b(\cdot)$ is measured by the following two metrics:
\begin{enumerate}
	\item \textit{complexity} $|b|$, defined as the number of conditions in $b$;
	\item \textit{support} $|\mathcal{S}_{\mathcal{N}}(b)|$, i.e., the size of the support set $\mathcal{S}_{\mathcal{N}}(b)$, which is defined as the set (or index set) of observations in $\mathcal{N}$ that satisfy clause $b$, namely $\mathcal{S}_{\mathcal{N}}(b) = \{i\in\mathcal{N}: b(x_i)=1\}$.
\end{enumerate}
As convention, the term {\em support} may refer to the set $\mathcal{S}_{\mathcal{N}}(b)$ or the size of the set $|\mathcal{S}_{\mathcal{N}}(b)|$.
Furthermore, for convenience, when using a set notation in this paper, the index set or the original set may be referred to, where confusions are not expected.
\subsection{Minimizing Complexity and Maximizing Support}\label{sec.MC_MS}
The problems of minimizing complexity (MC) and maximizing support (MS) of the globally-consistent summary-explanation $b(\cdot)\to y_e$ were introduced in~\cite{rudinGloballyConsistentRuleBasedSummaryExplanations2019}.
The objective of the MC problem is to find $b$ with minimal complexity $|b|$, which can be formulated as the following integer programming (IP) model:
\begin{alignat}{3}
	\min_{b}  & \sum_{p\in P^e}b_p \label{eq.MC_s}\\
	\text{s.t.} & \sum_{p\in P^e}b_p(1-\delta_{i,p})\geq 1, \quad \forall i\in \mathcal{N}\setminus \mathcal{N}^e\label{eq.consistency}\\
				   & b_p\in\{0,1\}, \quad \forall p\in \mathcal{P}^e\label{eq.MC_e}
\end{alignat}
where the binary decision variable $b_p=1$ indicates that feature $p$ is presented in the resulting clause $b=\land_{p\in\mathcal{P}'}F_p(\cdot)$, i.e., $p\in \mathcal{P}'$; and $b_p=0$ otherwise.
Therefore, $b_p$ are also called feature variables.
In addition, $\delta_{i,p}\in\{0,1\}$ denotes whether observation $i$ satisfies binary feature $p$, i.e., $\delta_{i,p} = F_p(x_i)$; and $\mathcal{P}^e$ is the subset of features satisfied by observation $e$, i.e., $\mathcal{P}^e=\{p\in\mathcal{P}: \delta_{e,p}=1\}$.
Furthermore, $\mathcal{N}^e$ is the set of consistent observations, i.e., $\mathcal{N}^e=\{ i\in\mathcal{N}: y_i=y_e \}$; thus $\mathcal{N}\setminus \mathcal{N}^e$ is the non-consistent observation set where $y_i\neq y_e$ for all $i\in\mathcal{N}\setminus\mathcal{N}^e$.
The property of relevancy is ensured by selecting features in $P^e$ only; the property of consistency is ensured with (\ref{eq.consistency}) by making sure that for any observation with $y_i\neq y_e$, $b(x_i)=0$.
The solution of the MC model is fast, due to simplicity; however, it does not guarantee a large support.

The objective of the MS problem is to find $b$ with maximal support $|s_N(b)|$, which can be formulated as the following more complicated IP model:
\begin{alignat}{3}
	\max_{b, r}  & \sum_{i\in \mathcal{N}^e} r_i~~~~~~~\label{eq.MS_s} \\
	\textbf{s.t.}~ & \sum_{p\in P^e}b_p(1-\delta_{i,p})\geq 1, \quad\forall i\in \mathcal{N}\setminus \mathcal{N}^e\label{eq.consistency2}\\
	& \sum_{p\in P^e}b_p(1-\delta_{i,p}) \leq M(1-r_i), \quad\forall i\in \mathcal{N}^e\label{eq.support}\\
	& \sum_{p\in P^e}b_p\leq {M_c},\label{eq.complexity_M_c}\\
	& b_p, r_i\in\{0,1\}, \quad \forall p\in \mathcal{P}^e,\forall i\in \mathcal{N}^e\label{eq.MS_e}
\end{alignat}

where the decision variable $r_i\in\{0,1\}$ indicates whether observation $i$ belongs to the support of clause $b(\cdot)$, i.e., $b(x_i)=1$ if and only if $r_i=1$.
In addition, $M_c>0$ is the maximum complexity of $b(\cdot)$ allowed in the solution, and $M$ is a large constant that should satisfy $M\geq M_c$.
As in~\cite{rudinGloballyConsistentRuleBasedSummaryExplanations2019}, $M_c=4$ is used in this paper, which is a reasonable complexity of summary-explanation for credit evaluation.
Equation~(\ref{eq.support}) ensures that, for observation $i\in\mathcal{N}^e$ to be a support of $b(\cdot)$, all conditions in $b(\cdot)$ must be satisfied by observation $i$.
The support of the MS summary-explanation by solving (\ref{eq.MS_s})--(\ref{eq.MS_e}) is typically much greater than the MC summary-explanation by solving (\ref{eq.MC_s})--(\ref{eq.MC_e}); however, solving the MS model is also much slower, due to its complexity.

\subsection{The Third Metric: Consistency Level}\label{sec.MSqC}
A globally-consistent summary-explanation $b(\cdot)\rightarrow y$ can be viewed as a 1-consistent or 100\%-consistent rule, since it requires the property of consistency to hold true for all observations $i\in\mathcal{N}$.
However, finding such a 1-consistent rule can be difficult.
For a large practical dataset, both MC and MS models often result in rules with high complexity or small support, if not infeasible.
The reason is straightforward: it is highly possible that a 1-consistent rule with reasonable complexity (e.g., $M_c=4$) does not exist for a complicated dataset.
It is therefore natural to relax the 1-consistency to a lower consistency level, e.g., 0.9- or 0.8-consistency, which should be acceptable for summary-explanation in many practical domains, including credit evaluation.

Let us define $q$-consistency as the following property of a summary-explanation $b(\cdot)\rightarrow y$: for at least $q$ fraction of observations $i\in\mathcal{N}$, if $b(x_i)=1$, then $y_i=y$.
Furthermore, let $\mathcal{S}_\mathcal{N}(b,y)$ denote the set of consistent support, i.e., $\mathcal{S}_\mathcal{N}(b,y)=\{i\in\mathcal{N}:b(x_i)=1, y_i=y\}$.
Then, the \textit{consistency level} of a rule $b(\cdot)\to y$ can be formally defined as $c_\mathcal{N}(b, y)=|\mathcal{S}_\mathcal{N}(b,y)|/|\mathcal{S}_\mathcal{N}(b)|$.
Therefore, the $q$-consistency property is equivalent to $c_\mathcal{N}(b, y)\geq q$.

A $q$\textit{-consistent summary-explanation} $b(\cdot)\rightarrow y_e$ for an observation $(x_e,y_e)$, in a more readable language, can be expressed as: ``for over $q$ fraction of observations where $b(\cdot)$ is TRUE, the outcomes are $y_e$''.
As an example in credit evaluation, a $q$-consistent summary-explanation for an observation of the FICO dataset is as follows:
``for 594 people whose ExternalRiskEstimate $\leq 63$ and AverageMInFile $\leq 48$, over 86\% of them were predicted to default.''


The problem considered in this paper is to find the clause $b$ with maximal support, while subjecting to the $q$-consistency constraint $c_\mathcal{N}(b, y)\geq q$, referred to as the MSqC problem.
The MSqC problem can be formulated by extending the MS IP model (\ref{eq.MS_s})--(\ref{eq.MS_e}), as follows: 
\begin{alignat}{3}
	\max_{b, r}  & \sum_{i\in \mathcal{N}} r_i~~~~~~~\label{eq.MSqC_s} \\
	\textbf{s.t.}~& \sum_{p\in P^e}b_p(1-\delta_{i,p}) + r_i \geq 1, \quad\forall i\in \mathcal{N}\setminus \mathcal{N}^e\label{eq.consistency3}\\
	& \sum_{p\in P^e}b_p(1-\delta_{i,p}) \leq M(1-r_i), \quad\forall i\in \mathcal{N}\label{eq.support3}\\
	& \sum_{p\in P^e}b_p \leq {M_c}\label{eq.complexity_M_c3},\\
	& \sum_{i\in N}(a_i - q)r_i \geq 0,\label{eq.q-consistency}\\
	& b_p, r_i\in\{0,1\}, \quad\forall p\in \mathcal{P}^e,\forall i\in \mathcal{N}.\label{eq.MSqC_e}
\end{alignat}

Specifically, compared with the MS model (\ref{eq.MS_s})--(\ref{eq.MS_e}), four modifications are made: i) binary variables $r_i$ (indicating supportive observations) are introduced for all $i\in\mathcal{N}$, instead of only $\mathcal{N}^e$; ii) constraints (\ref{eq.consistency3}) now include $r_i$ to allow inconsistent support ($r_i=1$) for $i\in \mathcal{N}\setminus \mathcal{N}^e$; iii) constraints~(\ref{eq.support3}) now cover all $i\in\mathcal{N}$ instead of only $\mathcal{N}^e$; iv) Eq.~(\ref{eq.q-consistency}) is the $q$-consistency constraint (equivalent to $c_\mathcal{N}(b,y)\geq q$), where binary constants $a_i=1$ if and only if $i\in\mathcal{N}^e$.

\begin{remark}\label{rm.MSefficiency}
It has been proved in~\cite{rudinGloballyConsistentRuleBasedSummaryExplanations2019} that the MC model is NP-hard, and
it can be easily observed that the MC (\ref{eq.MC_s})--(\ref{eq.MC_e}), MS (\ref{eq.MS_s})--(\ref{eq.MS_e}), and MSqC (\ref{eq.MSqC_s})--(\ref{eq.MSqC_e}) models have increasing complexities.
In practice, the time efficiency of solving the MC model is acceptable for the moderate dataset,
due to simplicity.
However, the MS model takes much longer time, and the time efficiency is even worse for MSqC, which has $|\mathcal{N}\setminus\mathcal{N}^e|$ additional variables and $|\mathcal{N}|+1$ more constraints than the MS model.
In~\cite{rudinGloballyConsistentRuleBasedSummaryExplanations2019}, the MC model was solved typically within 15 seconds, whereas a time limit of 60 seconds had to be set for solving the MS model.
\end{remark}

%% file: tex/040methodology.tex
\section{Methodology}\label{Methodology}
In Sections~\ref{sec.MC_MS} and~\ref{sec.MSqC}, the MC, MS, and MSqC problems have been formulated as different IP models (without confusions, also referred to as the MC, MS, and MSqC models, respectively) for summary-explanation of a target observation.
In the literature, the MC and MS models are solved using branch-and-bound solvers such as SCIP~\cite{GamrathEtal2020ZR}.
However, traditional singleton branch-and-bound methods on these IP models are hardly applicable for domains with large datasets. The reasons are two-folded:
\begin{enumerate}
	\item Branch-and-bound methods, when applied on the models, are either ineffective or inefficient. The MC model is simple and solving is fast, but branch-and-bound exact solutions have poor supports; on the contrary, the MS and MSqC model aim to maximize supports, but their solution times are unacceptable for practical purposes. These are verified by experiments in Section~\ref{Computer experiments}.
	\item Subset solution nullifies optimality. In domains like credit evaluations, the optimization models are most likely to be solved over only a subset of the observation dataset, due to practical reasons such as data efficiency, privacy, and distributed storage. It follows that performing branch-and-bound exact solution over the subset of observations may be unnecessary, if optimality cannot be extrapolated to the entire dataset.
\end{enumerate}
Therefore, the aim of our research is to develop a better solution method for the MSqC model, which on the one hand is more efficient than the traditional singleton branch-and-bound method, and on the other hand finds solutions, i.e., summary-explanation rules, that can extrapolate more effectively from local (subset) data to global data.
In this section, important elements of the SIS-based WCS optimization method are introduced respectively, e.g., the SIS values of feature variables, the WCS optimization framework, and the global prior injection mechanism.


\subsection{Simplified Increased Support}


The proposed method is related to the technique of column generation, an example of which was presented in~\cite{dashBooleanDecisionRules2018} to optimize DNF boolean rules for global explanation.
In~\cite{dashBooleanDecisionRules2018}, the master IP has exponentially many variables, thus the column generation method was used, where the optimal solution is searched by iteratively updating and solving a smaller problem, called the Restricted Master LP (Restricted MLP), which is the LP relaxation of the master IP problem, using only a small collection of variables.
Specifically, in each iteration, a pricing problem is solved to identify the next variable with the lowest reduced cost (for minimization problem) to add to the collection.
Since the reduced cost is dependent on the variables (or dual variables) of the Restricted MLP, the collection of variables can only be updated iteratively, by solving the Restricted MLP and the pricing problem one after the other, which is inefficient, and can hardly be parallelized.

In column generation, the reduced cost of a variable specifies the maximum possible change of the objective function value per unit increase in that variable's value.
The term ``reduced cost'' indicates a minimization problem, where the variables with low reduced costs should be prioritized to be selected.
Since MS and MSqC are formulated as maximization of support, we use the term \textit{increased support} instead, and the variables with large increased support should be prioritized.
Here we propose to solve the MSqC problem by solving and merging the solutions of multiple smaller IPs, each derived from the MS problem using a subset of observation data and feature variables.
Each feature variable used in a smaller IP is drawn from the set of feature variables of the original MS problem according to its  \textit{simplified increased support} (SIS), which is defined as the increased support of the variable with all dependent dual variables dropped.
In the following, we demonstrate how the SIS of a feature variable $b_p$ is derived.
First, the Lagrangian function of the MS model is as follows:
{\setlength\abovedisplayskip{0.5cm}
\setlength\belowdisplayskip{0.5cm}
\begin{small}
\begin{equation}
\begin{aligned}
L(b,r,\mu,\lambda,\gamma)=&\sum_{i\in\mathcal{N}^e}r_i + \sum_{i\in \mathcal{N}\setminus\mathcal{N}^e}\sum_{p\in P^e}\mu_i b_p(1-\delta_{i,p}) \\&- \sum_{i\in \mathcal{N}\setminus\mathcal{N}^e}\mu_i
- \sum_{i\in \mathcal{N}^e}\sum_{p\in P^e}\lambda_i b_p(1-\delta_{i,p}) \\&+ M\sum_{i\in \mathcal{N}^e}\lambda_i (1-r_i)
+\gamma (M_c - \sum_{p\in\mathcal{P}^e}b_p)\\&\geq \sum_{i\in\mathcal{N}^e}r_i,
\end{aligned}
\end{equation}
\end{small}}
where $\mu_i\geq 0, i\in\mathcal{N}\setminus\mathcal{N}^e$ are the dual variables for constraints~(\ref{eq.consistency2}), $\lambda_i\geq 0, i\in\mathcal{N}^e$ are the dual variables for constraints~(\ref{eq.support}), and $\gamma\geq 0$ is the dual variable for constraint~(\ref{eq.complexity_M_c}).
Thus, the increased support of a binary feature variable $b_p$ is given by
{\setlength\abovedisplayskip{0.5cm}
\setlength\belowdisplayskip{0.5cm}
\begin{equation}\label{eq.increasedSupport_MS}
\begin{aligned}
	L_p(\mu,\lambda,\gamma) &= L|_{b_p=1}-L|_{b_p=0}\\
	&=\sum_{i\in\mathcal{N}\setminus\mathcal{N}^e}\mu_i (1-\delta_{i,p})\\& - \sum_{i\in\mathcal{N}^e} \lambda_i (1-\delta_{i,p})-\gamma.
\end{aligned}
\end{equation}}
Since $L$ is the upper bound of the support, the increase of the upper bound specifies the maximum possible increase of the support.
Then, by dropping (set to zero) the dual variables $\mu_i$, $\lambda_i$, and $\gamma$ in the increased support (\ref{eq.increasedSupport_MS}), the SIS of each feature $p$ is obtained as
{\setlength\abovedisplayskip{0.5cm}
\setlength\belowdisplayskip{0.5cm}
\begin{equation}\label{eq.MS_SIS}
s_p=L_p(0)=\sum_{i\in\mathcal{N}^e}\delta_{i,p} - \sum_{i\in\mathcal{N}\setminus\mathcal{N}^e}\delta_{i,p}.
\end{equation}}
Matching our intuition, this is to weight a feature $p$ by the frequency it is satisfied by observations with $y_i=y_p$, minus the frequency it is satisfied by observations with $y_i\neq y_p$.
Additionally, it can be seen that the SIS value does not rely on primal or dual variables, and thus can be determined before solution.
\subsection{Weighted Column Sampling Optimization}\label{sec.WCS}
\begin{figure}
\centering
\includegraphics[width=0.36\textwidth]{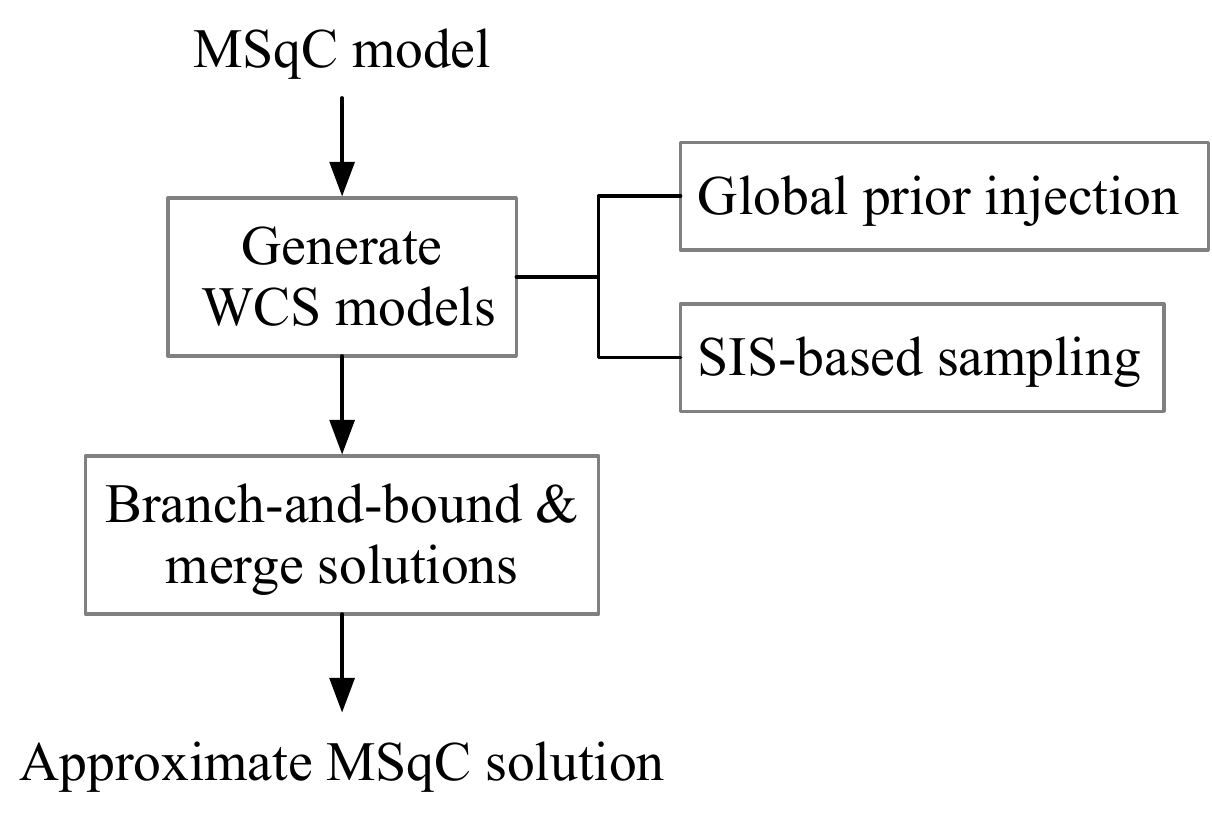}
\caption{Weighted column sampling (WCS) optimization} \label{fig.WCSframework}
\end{figure}

The proposed weighted column sampling (WCS) optimization framework can be divided into creating, solving, and merging the solutions of multiple smaller IPs, which are referred to as the WCS models.
Each WCS model $i$ is derived from the MS model (\ref{eq.MS_s})--(\ref{eq.MS_e}) by considering only a subset $\mathcal{N}_i\subseteq \mathcal{N}$ of observations, and using only a subset $\mathcal{P}_i^e\subseteq \mathcal{P}^e$ of feature variables $b_p$ that are sampled according to their SIS values~(\ref{eq.MS_SIS}). The overall procedure of the WCS optimization is shown in Figure~\ref{fig.WCSframework}, and the major steps are described as follows.
\begin{enumerate}
	\item Generate WCS models with SIS-based sampling. After computing the SIS values with (\ref{eq.MS_SIS}), each feature variable $b_p$ is sampled according the probability $\text{Prob}_p=\sigma_p(s')$ with
\begin{equation}\label{eq.prob_p}
	\sigma_p(x)=\frac{e^{x_p}}{\sum_ke^{x_k}}, \text{ and } s'_p = \frac{as_p}{\text{max}(s)},
\end{equation}
where $\sigma(\cdot)$ denotes the Softmax function, $s$ denotes the SIS vector~(\ref{eq.MS_SIS}), $s'$ is the normalized and scaled SIS with factor $a$, and $s'_p$ is the $p$th element of $s'$ for feature variable $b_p$.
In this step, $m$ WCS IPs are created, each by sampling $|\mathcal{P}'^e_i|=\rho |\mathcal{P}^e|$ (rounded, $0<\rho<1$) feature variables according to the probability $\text{Prob}_p$ (\ref{eq.prob_p}), and $|\mathcal{N}_i|=n_\text{wcs}$ observations uniformly from the observations set $\mathcal{N}$.
	\item Solve WCS models with multiple branch-and-bound and merge the results.
 The WCS models are IPs that can be solved via mainstream solvers, such as SCIP~\cite{GamrathEtal2020ZR}.
Finally, the WCS framework selects one of the optimal solutions of the WCS models that are feasible (after zero-padding) for MSqC (\ref{eq.MSqC_s})--(\ref{eq.MSqC_e}), and returns it as a sub-optimal solution of the MSqC problem.
It can be easily observed that any feasible solution $b$ of the WCS IPs that satisfies $c_\mathcal{N}(b,y)\geq q$ is also feasible for MSqC (\ref{eq.MSqC_s})--(\ref{eq.MSqC_e}).
Thus, we can simply select the solution with the largest support that satisfy $c_\mathcal{N}(b,y)\geq q$ as a sub-optimal solution for the MSqC problem (\ref{eq.MSqC_s})--(\ref{eq.MSqC_e}).
\end{enumerate}

\subsection{Global Prior Injection}\label{sec.approx_glob}

As mentioned earlier, in practice, the optimization models are most likely to be solved over only a subset of the observation dataset, due to reasons such as privacy and efficiency.
Therefore, besides time efficiency, it is also desirable to improve the extrapolation effectiveness from local data to global data.

Let $\mathcal{D}$ denote the index set of the global observation data (or simply called the global dataset), and $\mathcal{N}\subseteq \mathcal{D}$ denote the index set of a local observation dataset (or simply called the local dataset) that is drawn randomly from the global data and used to find a summary-explanation rule~$b\to y_e$ for a target observation~$(x_e, y_e), e\in \mathcal{N}$.
The local support~$|\mathcal{S}_\mathcal{N}(b)|$ and local consistency level~$c_\mathcal{N}(b,y_e)$ can be computed with a given solution, i.e., clause $b$, the target label $y_e$, and the local dataset $\mathcal{N}$.
However, the global support~$|\mathcal{S}_\mathcal{D}(b)|$ and global consistency level~$|c_\mathcal{D}(b,y)|$ need to be evaluated over the entire global dataset $\mathcal{D}$, which is unavailable during local solution of the optimization problem.
Therefore, the challenge in designing a solution method is improving global performance without accessing the global data during local solution.

The proposed WCS method meets this challenge by injecting global prior information into the local solution process of each WCS IP model using the global SIS $s_p^\text{gl}$, which is extended from (\ref{eq.MS_SIS}) as follows:
\begin{equation}\label{eq.MS_SIS_global}
	s_p^\text{gl}=\sum_{i\in\mathcal{D}^e}\delta_{i,p} - \sum_{i\in\mathcal{D}\setminus\mathcal{D}^e}\delta_{i,p},
\end{equation}
where $\mathcal{D}^e$ denotes the set of consistent observations in the global dataset $\mathcal{D}$, i.e., $\mathcal{D}^e=\{ i\in\mathcal{D}: y_i=y_e \}$.
Let $\sum_{i\in\mathcal{D}^e}\delta_{i,p}=\Delta_{p,y_e}$, and since $y_i$ are binary, we have $\sum_{i\in\mathcal{D}\setminus\mathcal{D}^e}\delta_{i,p}=\Delta_{p,1-y_e}$.
It can be observed that $\Delta_{p,0}$ and $\Delta_{p,1}$ only depend on the global dataset $\mathcal{D}$; therefore, they can be determined before solution and used for multiple summary-explanation requests.
\subsection{Branch-and-Bound Variants}
In Section~\ref{Problem formulation}, three IP optimization models have been introduced for summary-explanation, i.e., MC, MS, and MSqC.
These models are typically solved using branch-and-bound solvers, such as SCIP, where different tricks can be used to further improve the solution time efficiency. Here, four different branch-and-bound implementations are considered as follows.
\begin{enumerate}
	\item MC initialization. In~\cite{rudinGloballyConsistentRuleBasedSummaryExplanations2019}, the authors mentioned using MC's solution as an initial solution for the MS problem, which is referred to here as MC initialization (denoted as MCinit).
	\item SOS1 constraints. In all three models, there are sets of feature variables at most one of which can be non-zero; therefore, the special order sets of type 1 (SOS1) constraints can be used to narrow down the solution space.
	\item MCinit+SOS1. MC initialization and SOS1 constraints can be used together in one implementation, which is denoted as MCinit+SOS1.
	\item Vanilla implementations. Implementations without using SOS1 constraints or MC initialization are called ``vanilla'' implementations.
\end{enumerate}
Note that these varying implementations of branch-and-bound solution procedure only affect the time efficiencies in solving the models, and do not change the solution values.



\subsection{Feature Sampling Alternative}

The proposed SIS-based WCS optimization method samples feature variables according to their SIS values' associated Softmax probabilities~(\ref{eq.prob_p}).
As an alternative for comparison, the SIS-based feature sampling approach in WCS can be replaced with a uniform sampling approach, called random column sampling (RCS), which associates equal sampling probabilities for each feature variable.
The RCS method can be used to verify the effectiveness of the SIS-based feature sampling approach for WCS.

\subsection{Performance metrics}\label{sec.metrics}

As mentioned earlier, in practice, the optimization models are most likely to be solved over only a subset of the observation dataset, due to reasons such as privacy and efficiency.
To verify the extrapolation effectiveness from local data to global data, several extra performance metrics are needed in addition to solution time, namely, the local support~$|\mathcal{S}_\mathcal{N}(b)|$ and local consistency level~$c_\mathcal{N}(b,y_e)$ evaluated on the local dataset $\mathcal{N}$, and the global support~$|\mathcal{S}_\mathcal{D}(b)|$ and global consistency level~$c_\mathcal{D}(b,y)$ evaluated on the global dataset $\mathcal{D}$.

%% file: tex/050experiments.tex
\section{Computer experiments}\label{Computer experiments}
In this section, we first compare the different branch-and-bound implementations for the three models, i.e., MC (\ref{eq.MC_s})--(\ref{eq.MC_e}), MS (\ref{eq.MS_s})--(\ref{eq.MS_e}), and MSqC (\ref{eq.MSqC_s})--(\ref{eq.MSqC_e}).
For each model, the best implementation is used as the model's singleton branch-and-bound solution benchmark.
In addition, as a WCS alternative, the RCS method is also used as a solution benchmark.
Then, the proposed SIS-based WCS method is evaluated and compared against the traditional singleton branch-and-bound solution benchmarks, as well as the RCS method.

In the experiments, the FICO explainability dataset~\cite{ExplainableMachineLearning} is used. For each pair of solution method and size of the local dataset $|\mathcal{N}|=\alpha |\mathcal{D}|$, $0<\alpha\leq 1$, 40 solution runs are performed, each with a new re-sampled local dataset $\mathcal{N}$.
As convention, the metrics are computed by averaging over multiple solution runs using 1-shifted geometric mean, which has the advantage of being insensitive to both large outliers and small outliers~\cite{gasseExactCombinatorialOptimization2019, pengHeavyHeadSamplingFast2022}.
The experiments were conducted on a Mac Mini desktop computer with M1 chip.

\subsection{Solution Method Benchmarks}\label{sec.exact_sol_bm}

\begin{table*}
\centering
\caption{Average solution time (seconds) from 40 evaluations with random target observations $(x_e,y_e)$, $e\in\mathcal{N}$, computed with 1-shifted geometric mean. Model MSqC is solved with consistency level $q=0.85$. Solution time is capped at 2 hours. Bold font indicates each model's benchmark solution method.}\label{tab.variants_summary}
\begin{tabular}{|r|r|r|r|r|r|r|r|}
\hline
\multicolumn{1}{|l|}{~} & \multicolumn{2}{c|}{MC} & \multicolumn{4}{c|}{MS} & \multicolumn{1}{c|}{MSqC} \\
\hline
\multicolumn{1}{|c|}{$|\mathcal{N}|$} & \multicolumn{1}{c|}{\textbf{Vanilla}} & \multicolumn{1}{c|}{SOS1} & \multicolumn{1}{c|}{Vanilla} & \multicolumn{1}{c|}{\textbf{SOS1}} & \multicolumn{1}{c|}{Init} & \multicolumn{1}{c|}{InitSOS1} & \multicolumn{1}{c|}{\textbf{SOS1}} \\
\hline
0.1K & \textbf{0.0049} & 0.0172 & 3.2700 & \textbf{1.8505} & 3.1957 & 1.9709 & \textbf{5.5524} \\
\hline
0.4K & \textbf{0.0220} & 0.0699 & 42.8839 & \textbf{21.2645} & 43.1949 & 25.8114 & \textbf{252.2815} \\
\hline
0.7K & \textbf{0.0500} & 0.1244 & 88.2883 & \textbf{50.9096} & 95.1497 & 59.1170 & \textbf{1170.2276} \\
\hline
1K & \textbf{0.0643} & 0.1684 & 157.0540 & \textbf{100.6315} & 142.8052 & 98.0638 & \textbf{2374.9803}\\
\hline
4K & \textbf{0.3318} & 0.7905 & 1233.8075 & \textbf{557.5833} & 1117.0637 & 567.8364 & $\mathbf{> 7200}$\\
\hline
7K & \textbf{0.9006} & 1.4097 & $>7200$ & \textbf{1234.8625} & $>7200$ & 1248.3138 & $\mathbf{> 7200}$ \\
\hline
10K & \textbf{1.2297} & 1.9588 & $>7200$ & \textbf{1851.8545} & $>7200$ & 1796.3915 & $\mathbf{> 7200}$ \\
\hline
\end{tabular}
\end{table*}
The branch-and-bound methods are carried out with the SCIP solver~\cite{GamrathEtal2020ZR}, using the sampled local dataset $\mathcal{N}$ and a time limit of 2 hours.
Model MSqC is solved with consistency level $q=0.85$.
Table~\ref{tab.variants_summary} compares the time efficiencies of the different branch-and-bound variants of the MC and MS models.
As seen from Table~\ref{tab.variants_summary}, MC(Vanilla) and MS(SOS1) are the best branch-and-bound solution implementations for the MC and MS models, respectively.
The primary observations can be summarized as follows:
\begin{enumerate}
	\item MC initialization does not significantly improve the performance of the MS model, whether using SOS1 constraints or not.
	\item The SOS1 constraints slow down the solution of the MC model, while significantly accelerates the solution of the MS model. This can be reasoned as follows. On the one hand, including the SOS1 constraints narrows down the solution space; on the other hand, adding the SOS1 constraints increases the complexity of the model. For the simple MC model, the extra time for dealing with the SOS1 constraints dominates over the benefits of the smaller solution space. For the more complicated MS model, on the other hand, the smaller solution space through introducing the SOS1 constraints is dominant.
\end{enumerate}
Therefore, MC(Vanilla) and MS(SOS1) will be used as the benchmark singleton branch-and-bound methods for the MC and MS models, respectively.
Due to the similarity between the MS and MSqC models, the comparison of the different branch-and-bound variants for MSqC is omitted, and MSqC(SOS1) will be used as the benchmark branch-and-bound solution method for MSqC.
In addition, as a WCS alternative, the RCS method is also used as a solution benchmark.
As introduced in Section~\ref{sec.WCS}, the solution of the SIS-based WCS method is obtained by merging the solutions of $m$ small MS models, referred to as WCS IPs, each considering only a subset $\mathcal{N}_i\subseteq \mathcal{N}$ of observations (uniformly sampled) and a subset $\mathcal{P}_i^e\subseteq \mathcal{P}^e$ of feature variables (sampled according to the SIS values).
The RCS method is obtained by replacing the SIS-based sampling in WCS with uniform sampling, which is useful for evaluating the effectiveness of the SIS-based sampling in WCS.
In the experiments, $m=40$, $|\mathcal{N}_i|=n_\text{wcs}=100$, and $|\mathcal{P}_i^e|=|\mathcal{P}^e|/4$ are used.

\subsection{Evaluation of The Methods}
In the following, the different solution methods, i.e., the branch-and-bound benchmark methods, RCS, and the proposed WCS method, are compared, by studying their local solution effectiveness and global extrapolation effectiveness, respectively.
\subsubsection{Local Solution Effectiveness}


The local solution effectiveness of the different methods are studied by first comparing the time efficiencies, i.e., solution times, of the different methods, then the local supports and local consistencies of their solutions.

As shown in Tables~\ref{tab.variants_summary} and~\ref{tab.time_bm_wcs}, the solution times of the branch-and-bound solution methods are highly dependent on the local data size $|\mathcal{N}|$.
In contrast, the RCS and WCS methods are more scalable in terms of time efficiency.
It can be seen from Table~\ref{tab.time_bm_wcs} that the solution times of the RCS and the SIS-based WCS methods change little as $|\mathcal{N}|$ increases from 0.1K to 10K, while those of the branch-and-bound solution benchmarks increase drastically.

\begin{table}
\centering
\caption{Solution time of the proposed SIS-based WCS method and the benchmark solution methods, i.e., MC(Vanilla), MS(SOS1), MSqC(SOS1), and RCS.}\label{tab.time_bm_wcs}
\begin{tabular}{|r|r|r|r|r|r|}
\hline
\multicolumn{1}{|r|}{$|\mathcal{N}|$} & \multicolumn{1}{r|}{MC} & \multicolumn{1}{r|}{MS} & \multicolumn{1}{r|}{MSqC} & \multicolumn{1}{r|}{RCS} & \multicolumn{1}{r|}{WCS} \\
\hline
0.1K & 0.0049 & 1.8505 & 5.5524 & 3.8941 & 3.4727 \\
\hline
0.4K & 0.0220 & 21.2645 & 252.2816 & 5.1050 & 3.7230 \\
\hline
0.7K & 0.0500 & 50.9096 & 1170.2277 & 4.6023 & 3.9933 \\
\hline
1K & 0.0643 & 100.6315 & 2374.9803 & 5.0522 & 3.8300 \\
\hline
4K & 0.3318 & 557.5833 & $>7200$ & 5.0952 & 4.0497 \\
\hline
7K & 0.9006 & 1234.8625 & $>7200$ & 4.9440 & 3.8923 \\
\hline
10K & 1.2297 & 1851.8545 & $>7200$ & 5.1366 & 3.6883 \\
\hline
\end{tabular}
\end{table}



\begin{table}
\centering
\caption{Local support $|\mathcal{S}_\mathcal{N}(b)|$ and local consistency $c_\mathcal{N}(b,y_e)$ of the proposed SIS-based WCS method and the benchmark solution methods, i.e., MC(Vanilla), MS(SOS1), MSqC(SOS1), and RCS.}\label{tab.wcs_local}
{
\begin{tabular}{|r|r|r|r|r|r|r|r|r|}
\hline
~ & \multicolumn{1}{c|}{MC} & \multicolumn{1}{c|}{MS} & \multicolumn{2}{c|}{MSqC} & \multicolumn{2}{c|}{RCS} & \multicolumn{2}{c|}{WCS} \\
\hline
$|\mathcal{N}|$ & \multicolumn{1}{c|}{$|\mathcal{S}_\mathcal{N}|$} & \multicolumn{1}{c|}{$|\mathcal{S}_\mathcal{N}|$} & \multicolumn{1}{c|}{$|\mathcal{S}_\mathcal{N}|$} & \multicolumn{1}{c|}{$c_\mathcal{N}$} & \multicolumn{1}{c|}{$|\mathcal{S}_\mathcal{N}|$} & \multicolumn{1}{c|}{$c_\mathcal{N}$} & \multicolumn{1}{c|}{$|\mathcal{S}_\mathcal{N}|$} & \multicolumn{1}{c|}{$c_\mathcal{N}$} \\
\hline
0.1K & 2.16 & 18.10 & \textbf{35.49} & 0.86 & 16.73 & 1.00 & 16.74 & 1.00 \\
\hline
0.4K & 3.26 & 28.80 & \textbf{93.59} & 0.86 & 61.71 & 0.87 & 66.78 & 0.87 \\
\hline
0.7K & 2.95 & 30.11 & \textbf{130.81} & 0.85 & 74.88 & 0.86 & 94.17 & 0.83 \\
\hline
1K & 3.49 & 30.07 & \textbf{155.97} & 0.85 & 85.31 & 0.74 & 127.67 & 0.84 \\
\hline
4K & 4.09 & 26.41 & N/A & N/A & 502.57 & 0.79 & \textbf{541.22} & 0.83 \\
\hline
7K & 4.79 & 22.01 & N/A & N/A & 819.51 & 0.79 & \textbf{900.49} & 0.79 \\
\hline
10K & 5.17 & 21.54 & N/A & N/A & 897.96 & 0.76 & \textbf{961.64} & 0.77 \\
\hline
\end{tabular}}
\end{table}
Table~\ref{tab.wcs_local} shows the local supports and local consistencies of the different methods.
The local consistencies of the MC and MS solutions are omitted, since the MC and MS models guarantees 1-consistency inherently.
It can be seen that MSqC has the largest local support, since it does the branch-and-bound solution of the maximization of local support, subjecting to consistency level $c_\mathcal{N}(b,y_e)\geq 0.85$. However, when data size $|\mathcal{N}|\geq 4$K, it takes too long for MSqC(SOS1) to complete solution within the 2-hour time limit (thus N/A in Table~\ref{tab.wcs_local}).
The other extreme is MC(Vanilla), which has the best time efficiency but worst support.
In contrast, the SIS-based WCS method finds solutions with the second largest local supports, while also achieving the second best time efficiency.
It is worth pointing out that the scalabilities of RCS and WCS are due to the use of fixed sub-problem data sizes $|\mathcal{N}|_i=n_\text{wcs}$, which sacrifices local consistencies of the solution as data size increases.
Therefore, as $|\mathcal{N}|$ increases further (greater than 10K), a greater sub-problem data size $n_\text{wcs}$ should be used to improve solution quality, which will then increases solution time.

\begin{figure*}
	\centering
	\includegraphics[width=.8\linewidth]{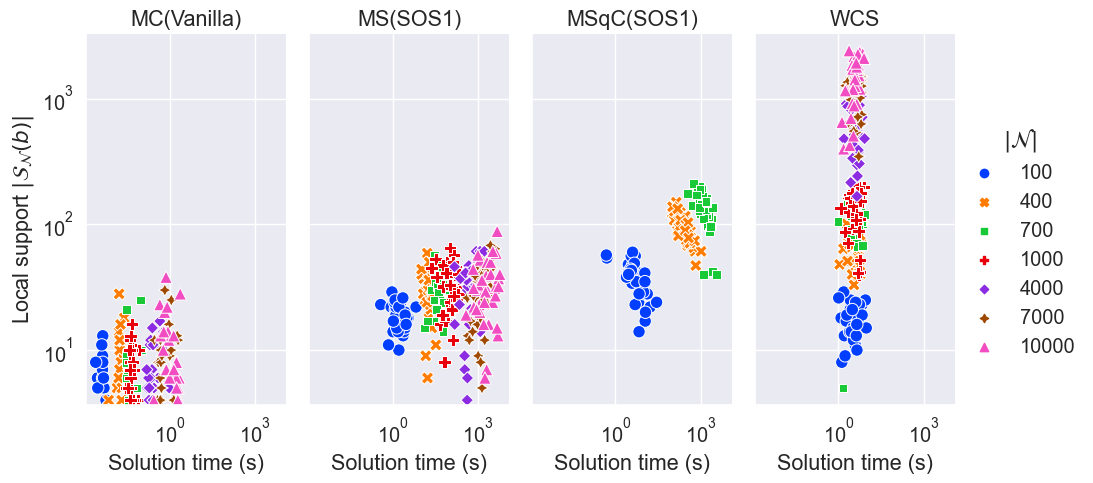}
	\caption{Local support vs. solution time of the methods, i.e., MC(Vanilla), MS(SOS1), MSqC(SOS1), and the proposed SIS-based WCS method. Model MSqC is solved with consistency level $q=0.85$. For each local data size $|\mathcal{N}|$, each method is run 40 times with resampled local dataset $\mathcal{N}$.} \label{fig.scatterplots_innerSup_vs_time}
\end{figure*}

It can also be seen from Tables~\ref{tab.time_bm_wcs} and~\ref{tab.wcs_local} that the SIS-based WCS method surpasses RCS in all three aspects, i.e., solution efficiency, local support, and local consistency.
Therefore, the RCS method is omitted in Fig.~\ref{fig.scatterplots_innerSup_vs_time}, which shows the solution times and local supports of solutions of the different methods.
It can be observed clearly from Fig.~\ref{fig.scatterplots_innerSup_vs_time} that the WCS method's time efficiency scales well with the local data size $|\mathcal{N}|$, and finds solutions with much greater local support than the other methods when $|\mathcal{N}|$ is large.
\subsubsection{Global Extrapolation Effectiveness}

\begin{figure*}
	\centering
	\includegraphics[width=.8\linewidth]{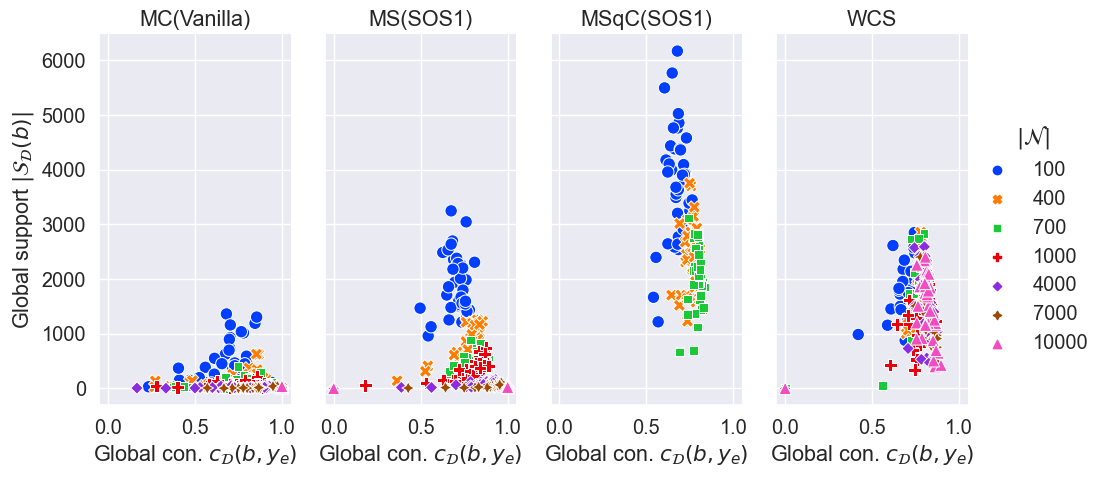}
	\caption{Global support vs. global consistency of the solutions of the methods, i.e., MC(Vanilla), MS(SOS1), MSqC(SOS1), and the proposed SIS-based WCS method. Model MSqC is solved with consistency level $q=0.85$. For each local data size $|\mathcal{N}|$, each method is run 40 times with resampled local dataset $\mathcal{N}$.} \label{fig.scatterplots_outerSup_vs_OuterCons}
\end{figure*}

As mentioned earlier, in practice, the models are most likely to be solved over only a subset of the observation dataset, due to reasons such as privacy and efficiency.
Therefore, it is desirable to evaluate the effectiveness of extrapolation from local data to global data.

As shown in Table~\ref{tab.wcs_global}, when extrapolating to the global dataset $\mathcal{D}$, the solutions of MSqC still have the greatest global supports.
However, considering its time efficiency, MSqC is not appropriate for practical use.
In contrast, the proposed SIS-based WCS method not only has the second best time efficiency (Table~\ref{tab.time_bm_wcs}), but also finds solutions with large global supports and large global consistencies (See Table~\ref{tab.wcs_global} and Fig.~\ref{fig.scatterplots_outerSup_vs_OuterCons}).
The advantage of WCS on global consistency is more apparent when $|\mathcal{N}|\leq 0.4|\mathcal{D}|=4K$, since the impact of global extrapolation is stronger for smaller local-to-global ratio.
The advantage of the WCS method on global extrapolation is due to the global prior injection through global SIS $s_p^\text{gl}$, as mentioned in Section~\ref{sec.approx_glob}.
\begin{table}
\centering
\caption{Global support $|\mathcal{S}_\mathcal{D}(b)|$ and global consistency $c_\mathcal{D}(b,y_e)$ of the proposed SIS-based WCS method and the benchmark solution methods, i.e., MC(Vanilla), MS(SOS1), MSqC(SOS1), and RCS.}\label{tab.wcs_global}
{
\begin{tabular}{|r|r|r|r|r|r|}
\multicolumn{6}{c}{(a) Global support $|\mathcal{S}_\mathcal{D}(b)|$} \\
\hline
\multicolumn{1}{|c|}{relax} & \multicolumn{1}{c|}{MC} & \multicolumn{1}{c|}{MS} & \multicolumn{1}{c|}{MSqC} & \multicolumn{1}{c|}{RCS} & \multicolumn{1}{c|}{WCS} \\
\hline
0.1K & 333.2048 & 1765.2343 & \textbf{3546.0163} & 1788.2255 & 1693.7984 \\
\hline
0.4K & 101.5732 & 721.2592 & \textbf{2311.3068} & 1558.0619 & 1666.5095 \\
\hline
0.7K & 62.4926 & 439.2204 & \textbf{1865.6976} & 1052.9064 & 1275.6396 \\
\hline
1K & 46.4917 & 321.7781 & \textbf{1543.2629} & 683.8612 & 1266.3384 \\
\hline
4K & 13.1769 & 68.0760 & N/A & 1230.8258 & \textbf{1359.7509} \\
\hline
7K & 7.2883 & 33.0460 & N/A & 1164.7167 & \textbf{1270.4606} \\
\hline
10K & 5.1704 & 21.5444 & N/A & 897.9617 & \textbf{961.6415} \\
\hline
\multicolumn{6}{c}{} \\
\multicolumn{6}{c}{(b) Global consistency $c_\mathcal{D}(b,y_e)$} \\
\hline
\multicolumn{1}{|c|}{relax} & \multicolumn{1}{c|}{MC} & \multicolumn{1}{c|}{MS} & \multicolumn{1}{c|}{MSqC} & \multicolumn{1}{c|}{RCS} & \multicolumn{1}{c|}{WCS} \\
\hline
0.1K & 0.6828 & 0.7072 & 0.6728 & 0.6999 & 0.7191 \\
\hline
0.4K & 0.6984 & 0.7697 & 0.7596 & 0.7680 & 0.7849 \\
\hline
0.7K & 0.7232 & 0.7921 & 0.7866 & 0.7850 & 0.7576 \\
\hline
1K & 0.7519 & 0.7872 & 0.7800 & 0.6969 & 0.7853 \\
\hline
4K & 0.7895 & 0.8186 & N/A & 0.7869 & 0.8160 \\
\hline
7K & 0.8872 & 0.8495 & N/A & 0.7870 & 0.7904 \\
\hline
10K & 1.0000 & 1.0000 & N/A & 0.7618 & 0.7650 \\
\hline
\end{tabular}}
\end{table}

\subsubsection{Overall Performance}

The overall performance of the WCS method is clearer when looking at its rankings among all five methods, as shown in Table~\ref{tab.ranking}.
In summary, the WCS method is more applicable in practical scenarios, which simultaneously has decent time efficiency and large local/global supports, and also has large global consistency for global extrapolation when the local-to-global ratio is small enough, i.e., $|\mathcal{N}|/|\mathcal{D}|\leq 0.4$ ($|\mathcal{D}|=10$K in the experiment).

\begin{table}
\centering
\caption{Rankings of the proposed WCS method among all five methods, i.e., MC(Vanilla), MS(SOS1), MSqC(SOS1), RCS, and WCS, in terms of different metrics. Bold font indicates the "decent" ranks.}\label{tab.ranking}
{
\begin{tabular}{|l|r|r|r|r|r|r|r|}
\hline
Local data size $|\mathcal{N}|$ & 0.1K & 0.4K & 0.7K & 1K & 4K & 7K & 10K \\
\hline
Solution time & 3 & \textbf{2} & \textbf{2} & \textbf{2} & \textbf{2} & \textbf{2} & \textbf{2} \\
\hline
Local Support & 3 & \textbf{2} & \textbf{2} & \textbf{2} & \textbf{1} & \textbf{1} & \textbf{1} \\
\hline
Global Support & 4 & \textbf{2} & \textbf{2} & \textbf{2} & \textbf{1} & \textbf{1} & \textbf{1} \\
\hline
Global Consistency & \textbf{1} & \textbf{1} & 4 & \textbf{2} & \textbf{2} & 3 & 3 \\
\hline
\end{tabular}}
\end{table}

%% file: tex/060conclusion.tex
\section{Conclusions}\label{conclusions}
In this paper, we have argued that asking summary-explanations to have 100\%-consistency is most likely an overkill in practical scenarios, and that summary-explanations for practical large datasets are natural to have small inconsistencies, which can be used to effectively increase their supports.
Motivated by this, in this paper, the MSqC problem has been formulated, which aims for greater support at the cost of slightly lower consistency.
Then, it has been noted that the MSqC problem is much more complicated than the MS problem, and that solving MSqC with branch-and-bound solvers for exact solutions are unrealistic in practice.
Therefore, the SIS-based WCS method has been proposed, which is much more scalable in efficiency for MSqC.
Furthermore, the global prior injection technique has been proposed and integrated in the SIS-based WCS method to help finding solutions with better global extrapolation effectiveness.
It has been verified in experiments that the SIS-based WCS method on MSqC is not only much faster, but also finds solutions with both greater supports and better global extrapolation effectivenesses.

%% file: tex/003acknowledgement.tex
\section{Acknowledgments}
This work was supported by the Natural Science Foundation of China under Grant 62006095, by the Natural Science Foundation of Hunan Province, China, under Grant 2021JJ40441, and by the Research Foundation of Education Bureau of Hunan Province, China, under Grants 20B470 and 20A407.

